%% file: main.tex
\documentclass[letter, 10pt, conference]{ieeeconf}      

\IEEEoverridecommandlockouts                              %
\usepackage{array,multicol,multirow,booktabs,longtable,tabularx}
\usepackage{subfig}
\usepackage{amssymb,mathtools,dsfont,bbm,cuted,stmaryrd}
\usepackage{xspace,algorithmic,algorithm}
 
\usepackage{amsthm}
\usepackage{booktabs}

\usepackage{tikz}
\usetikzlibrary{math,arrows.meta,spy,automata,intersections,decorations,arrows,decorations.markings,decorations.footprints,fadings,calc,trees,mindmap,shadows,decorations.text,patterns,positioning,shapes,patterns,decorations.pathmorphing,patterns.meta,fit,3d,plotmarks,svg.path}

\usepackage{pgfplots}

\usepgfplotslibrary{patchplots}

\usepackage{tikz-3dplot}
\usepackage{tkz-euclide}

\input{./tikz_styles}

\usepackage{placeins}
\usepackage{cuted}

\usepackage{flushend}

\let\labelindent\relax
\usepackage{enumitem}

\newcommand{\be}{\begin{equation}}
\newcommand{\ee}{\end{equation}}

\def\bal#1\eal{\begin{align}#1\end{align}}
\def\baln#1\ealn{\begin{align*}#1\end{align*}}

\newcommand{\ben}{\begin{equation*}}
\newcommand{\een}{\end{equation*}}

\newcommand{\bbm}{\begin{bmatrix}}
\newcommand{\ebm}{\end{bmatrix}}

\newcommand{\bBm}{\begin{Bmatrix}}
\newcommand{\eBm}{\end{Bmatrix}}

\newcommand{\bvm}{\begin{vmatrix}}
\newcommand{\evm}{\end{vmatrix}}

\newcommand{\bVm}{\begin{Vmatrix}}
\newcommand{\eVm}{\end{Vmatrix}}

\newcommand{\bpm}{\begin{pmatrix}}
\newcommand{\epm}{\end{pmatrix}}

\newcommand{\bnm}{\begin{matrix}}
\newcommand{\enm}{\end{matrix}}

\newcommand{\bi}{\begin{itemize}}
\newcommand{\ei}{\end{itemize}}

\newcommand{\bse}{\begin{subequations}}
\newcommand{\ese}{\end{subequations}}

\newenvironment{proof-sketch}{\noindent{ \textit{Sketch of Proof}:}\hspace*{0.5em}}

\theoremstyle{plain}

\newtheorem{prop}{Proposition}

\newtheorem{lem}{Lemma}

\newtheorem{rem}{Remark}

\newcommand{\includeFGO}[2][0.75]{\begin{tikzpicture}[baseline=-2pt]\node[#2, scale=#1] at(0,1pt){};\end{tikzpicture}}                                    %

\title{\LARGE \bf Real-time loosely coupled GNSS and IMU integration \\ via Factor Graph Optimization$^*$}

\author{Radu-Andrei Cioac\u a$^{1}$, Cristian Rusu$^{1}$, Paul Irofti$^{1}$,\\ Gianluca Caparra$^{2}$, Andrei-Alexandru Marinache$^{3}$, Florin Stoican$^{1}$
\thanks{$^{1}$R.-A. Cioac\u a, P. Irofti, C. Rusu and F. Stoican are with Three Tensors S.R.L., Romania {\small \{radu.cioaca,paul.irofti,}
        {\small cristian.rusu,florin.stoican\}@three-tensors.com}}%
\thanks{$^{2}$G. Caparra is with Navigation Systems Definition Section (TEC-SEN), European Space Agency.
        {\small Gianluca.Caparra@esa.int}}%
\thanks{$^{3}$A.-A. Marinache is with Romanian InSpace Engineering S.R.L. (RISE), Romania.
        {\small alexandru.marinache@roinspace.com}}%
\thanks{* The work is carried out under ESA NAVISP Element 1, which is devoted to the development of innovative PNT, systems, technologies, algorithms and techniques. Three of the authors are supported in part by the project “Romanian Hub for Artificial Intelligence - HRIA”, Smart Growth, Digitization and Financial Instruments Program, 2021-2027, MySMIS no. 351416.}%
}

\usepackage{graphicx}      

\usepackage{amsmath,amsfonts,amssymb,amsthm}
\usepackage[svgnames]{xcolor}

\usepackage[algo2e, linesnumbered, algoruled]{algorithm2e}

\usepackage{hyperref}

\begin{document}

\maketitle

\begin{abstract}
Accurate positioning, navigation, and timing (PNT) are fundamental to the operation of modern technologies and a key enabler of autonomous systems. A very important component of PNT is the Global Navigation Satellite System (GNSS) which ensures outdoor positioning. Modern research directions have pushed the performance of GNSS localization to new heights by fusing GNSS measurements with other sensory information, mainly measurements from Inertial Measurement Units (IMU). In this paper, we propose a loosely coupled architecture to integrate GNSS and IMU measurements using a Factor Graph Optimization (FGO) framework. Because the FGO method can be computationally challenging and often used as a post-processing method, our focus is on assessing its localization accuracy and service availability while operating in real-time in challenging environments (urban canyons). Experimental results on the UrbanNav-HK-MediumUrban-1 dataset show that the proposed approach achieves real-time operation and increased service availability compared to batch FGO methods. While this improvement comes at the cost of reduced positioning accuracy, the paper provides a detailed analysis of the trade-offs between accuracy, availability, and computational efficiency that characterize real-time FGO-based GNSS/IMU fusion.


\end{abstract}



\section{Introduction}

Positioning, Navigation, and Timing (PNT) \cite{doi:https://doi.org/10.1002/9781119458449.fmatter} technologies are essential for a large number of modern services and applications. For example, in outdoor navigation, the Global Navigation Satellite System (GNSS) is the main source of localization. Providing a reliable, statistically bounded position is of utmost importance in many scenarios, including for autonomous vehicles performing critical safety operations (autonomous taxis, drones, public transport, etc.). 

In urban environments, GNSS often suffers from outage, multipath, loss of Line-of-Sight, and interference (including jamming). To mitigate these issues, Inertial Measurement Unit (IMU) measurements are integrated with GNSS to improve the positioning accuracy. Similarly, IMU measurements themselves suffer from cumulative drift, leading to significant degradation of accuracy over time. As such, sensor fusion of GNSS and IMU data has been widely adopted in the literature. Classic approaches include the use of weighted least-squares estimators (WLS) and various variants of the Kalman Filter (KF). Lately, a flexible new framework called Factor Graph Optimization (FGO) has also been applied to the GNSS/IMU fusion problem \cite{wen_factor_2021}.

Factor Graphs, as proposed by \cite{10.1109/18.910572} in the statistical estimation literature, offer a mathematical framework to describe an optimization problem as a factorized model where measurements and states are connected in a graph-like fashion. FGO inherently facilitates the split of the optimization problem into smaller chunks that can later be combined to model the overall evolving estimation problem. Specialized optimizers then take advantage of the graph structure to efficiently perform multi-iteration operations. Historically, the framework has been applied with great success in robotics \cite{doi:10.1177/0278364906072768, doi:10.1177/0278364906065387}. 

Early applications of FGO to positioning focused on improving GNSS-only solutions. Examples include SLAM-based approaches~\cite{sunderhauf_towards_2012} and switchable constraints with incremental smoothing~\cite{sunderhauf_switchable_2013} for mitigating GNSS Non-Line-of-Sight (NLOS) and multipath effects. FGO has also enabled precise smartphone positioning~\cite{suzuki_precise_2023}, effectively reducing the impact of missing and noisy GNSS data.

More recent work has extended FGO to GNSS/IMU fusion. In tightly coupled architectures, raw GNSS measurements (pseudorange, Doppler, carrier phase, etc.) are fused directly with IMU data, forming fully integrated state estimates; notable examples and results are reported in~\cite{Taylornavi.653,WU202225}. When raw measurements are unavailable or simpler designs are preferred, loosely coupled systems combine GNSS-processed states (position, velocity, time) with IMU data. FGO was first applied in this context in~\cite{indelman_factor_2012}, outperforming both full nonlinear batch estimation and the Extended Kalman Filter (EKF). Subsequent work has repeatedly confirmed these benefits: \cite{wen_factor_2021} showed that FGO better exploits temporal correlations—particularly under non-Gaussian noise—while \cite{rs15215144} demonstrated that PPP-B2b/INS integration using FGO achieves at least a 15\% accuracy improvement over EKF in challenging urban environments. More recently, \cite{11028340} proposed a robust loosely coupled GNSS/IMU fusion scheme based on a Huber loss.

In this paper, we propose a loosely coupled solution to integrate GNSS and IMU measurements using FGO. Our study focuses on the robustness of the navigation solution and the ability to compute it in real-time. We evaluate the performance of our approach using well-known real-world datasets (focused on real-world urban driving scenarios) where GNSS signals are prone to outages or degradation. For the numerical computation of the results, we make use of the RTKLIB \cite{RTKLIB} and GTSAM \cite{gtsam} software libraries, while the well-known, publicly available UrbanNav dataset serves as the basis of the findings~\cite{Hsunavi.602}. 
We show numerical evidence {illustrating the behavior of FGO under real-time constraints and we quantify the trade-offs between accuracy, service availability, and computational load introduced by different configurations -- smoothing latency, marginalization lag, and IMU-only propagation.}


\section{Preliminaries}

\subsection{Factor Graph Optimization}

A Factor Graph is a probabilistic graphical model composed of two main objects: i) the nodes $x_k$ which represent the $k^\text{th}$ state vectors to be estimated and ii) the factors $f_i(\cdot)$ and $h_i(\cdot)$ which encode the dynamics of the system that relate the state vectors between themselves and to the measurements, respectively. Given $n$ measurements, we define the overall state of the graph to be $\mathcal{X} \in \{ x_0, x_1, \dots, x_n \}$. Therefore, the overall estimation problem, written as an optimization problem, is:
\begin{equation}
	\begin{aligned}
    J(\mathcal{X}) = \left(  \| x_0 - \bar{x}_0  \|_{\Sigma_0}^2  \right. &  \left. + \sum_{k=1}^n \| h_k(x_k) - z_k  \|_{\Sigma_R^{k}}^2 \right. \\
	&  \left. + \sum_{k=1}^n \| f_k(x_{k-1}) - x_k  \|_{\Sigma_Q^k}^2  \right).
	\end{aligned}
	\label{eq:FGO}
\end{equation}
In this formulation, $\bar{x}_0$ denotes the given initial state, $z_i$ are the recorded measurements, and $\Sigma_R^{k}$ and $\Sigma_Q^{k}$ are the associated covariance matrices. We use the Mahalanobis distance $\|r\|_{\Sigma}^2 = r^\top \Sigma^{-1} r$. Importantly, the optimization is performed over the entire state trajectory $\mathcal{X}$, not only the most recent state $x_n$. Both dynamics and measurements are allowed to be non-linear functions and, in general, the optimization horizon $n$ can be as large as computational resources allow. Note that as new measurements become available, all state vectors are updated, thereby improving past errors. Non-linearities are addressed at each optimization step by using linearized approximations of $f_i(\cdot)$ and $h_i(\cdot)$. The optimization problem in \eqref{eq:FGO} is solved via iterative methods, including well-established Gauss-Newton or Levenberg-Marquardt techniques or incremental smoothing methods such as iSAM2~\cite{isam2}.


While the FGO framework has many advantageous properties, there are some drawbacks. An important issue is the size of the problem. As the estimation procedure accumulates states and $n$ grows large, computational constraints are the bottleneck in the calculation of the  solution to \eqref{eq:FGO}.

A powerful strategy to reduce the size of the Factor Graph is called marginalization. The key point is that some of the state vectors can be fixed and therefore removed from the optimization graph, thus removing nodes from the graph and reducing the size of the optimization problem. In scenarios where the system naturally evolves in time, the main idea is to use a sliding window of a particular size, which excludes old measurements and states from the problem.




\subsection{GNSS and IMU equations}
A GNSS receiver computes its position by measuring how long it takes for radio signals, transmitted by several satellites, to reach its antenna. Each satellite continuously broadcasts a precisely time-stamped signal along with its orbital parameters. By correlating a local copy of the code with the received one, the receiver estimates the signal delay, which multiplied by the speed of light, gives an apparent distance to the satellite, known as the pseudorange \cite{teunissen_springer_2017}. The pseudorange can be modeled as
\begin{equation}
\rho^s = \|p_r - p^s\|_2
+ c(\delta t_r - \delta t^s)
+ T^s + I^s + \epsilon_\rho^s,
\end{equation}
where $p_r$ and $p^s$ denote the receiver and satellite `$s$' positions, $c$ is the speed of light, $\delta t_r$ and $\delta t^s$ are the receiver and satellite clock biases, $T^s$ and $I^s$ represent the tropospheric and ionospheric delays, and $\epsilon_\rho^s$ accounts for residual errors such as multipath and interference. By measuring pseudoranges to at least four satellites, the receiver can estimate its position and clock bias through trilateration. In urban environments, however, GNSS signals often suffer from NLOS reception and multipath effects, caused by reflections from surrounding structures. These phenomena distort the pseudoranges, leading to significant positioning errors or even loss of service when too few valid satellites are visible.

An IMU measures a platform’s motion by sensing specific force and angular rate along three orthogonal axes. Operating at high sampling rates (typically hundreds of hertz), it enables short-term motion estimation independent of external signals such as GNSS. The continuous-time IMU measurement equations are given by~\cite{groves_principles_2015}
\begin{equation}
\begin{aligned}
\tilde{f} &= R_{n}^{b}\!\left(a^{n} - g^{n}\right) + b_a + n_a,\\
\tilde{\omega} &= \omega_{ib}^{b} + b_g + n_g,
\end{aligned}
\end{equation}
where $\tilde{f}$ and $\tilde{\omega}$ are the measured specific force and angular rate, $R_{n}^{b}$ is the rotation from navigation to body frame, $a$ is the linear acceleration, $g$ is gravity, $\omega_{ib}^{b}$ is the true angular rate, and $b_a$, $b_g$ are the accelerometer and gyroscope biases with white noise terms $n_a$, $n_g$. These biases vary slowly over time due to temperature and mechanical effects and must be continuously estimated to prevent drift.

\begin{figure}[!ht]
    \centering
    \includegraphics[width=\columnwidth]{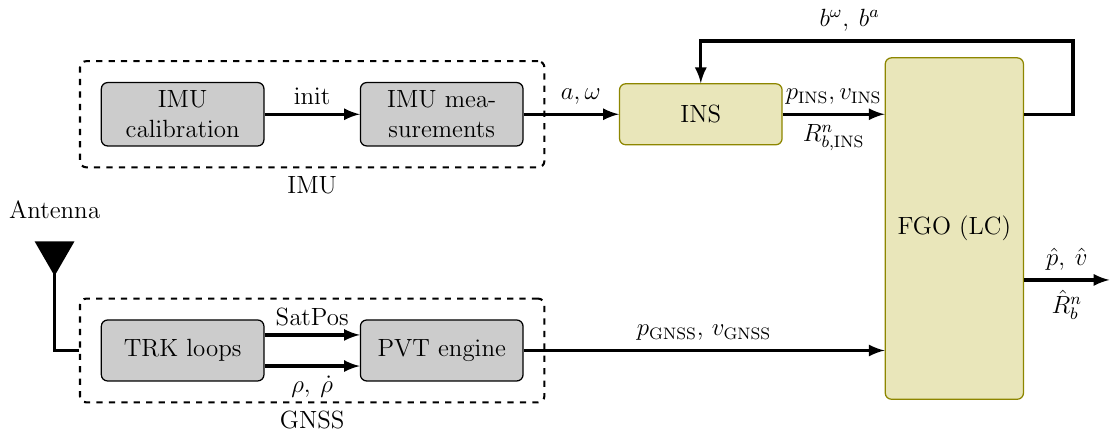}
    \caption{Loosely coupled GNSS/IMU integration scheme.}
    \label{fig:fgo-lc-scheme}
\end{figure}
By integrating accelerometer and gyroscope data, the IMU provides estimates of position, velocity, and attitude; 
\begin{figure*}[!htp]  
  \centering
  \includegraphics[width=\textwidth]{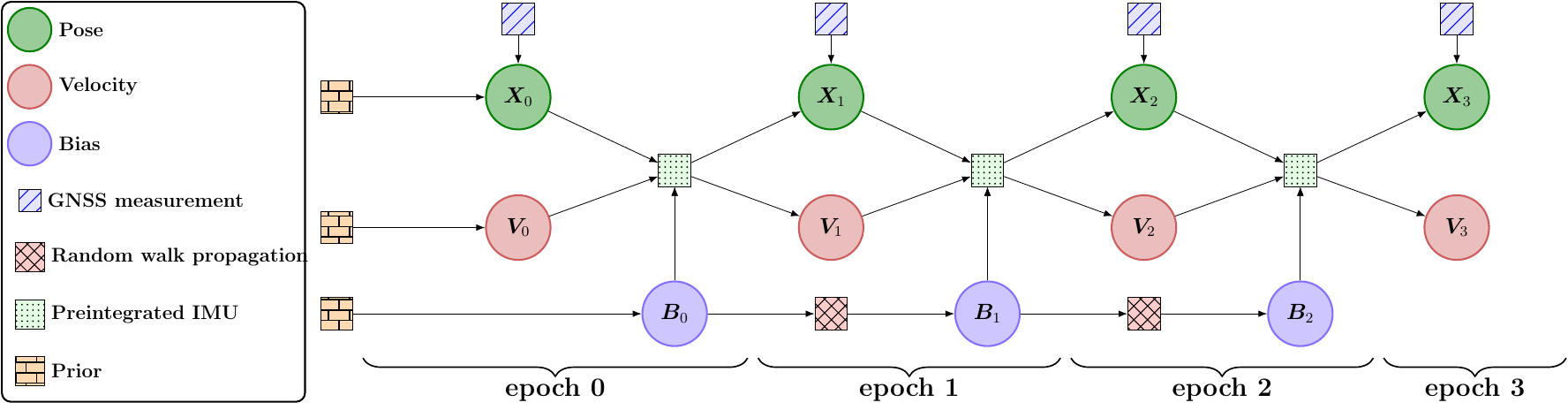}
  \caption{Loosely-coupled GNSS and IMU factor graph architecture.}
  \label{fig:fgo_scheme}
\end{figure*}

however, these estimates degrade quickly because integration amplifies measurement errors and bias drift.

Therefore, we propose and evaluate a loosely coupled GNSS/IMU architecture through an FGO scheme, as depicted in Fig.~\ref{fig:fgo-lc-scheme}.


\section{Loosely coupled factor graphs}

In this section we map the GNSS/IMU measurement fusion problem to the FGO framework. We discuss separately the nodes, factors and the FGO optimization problem.


\subsection{State vector}
Each node in the factor graph represents the navigation state at time~$k$, defined as
\begin{equation}
x_k = \{R_k,\, p_k,\, v_k,\, b_a^k,\, b_g^k\},
\end{equation}
where $R_k$ is the rotation matrix belonging to the special orthogonal group $\mathrm{SO}(3)$, representing the orientation from the body frame to the chosen navigation frame, $p_k$ is the position, $v_k$ the velocity, and $b_a^k$, $b_g^k$ the accelerometer and gyroscope biases.
The state vector is described inside the factor graph by distributing its components into variable nodes: (i) pose (\includeFGO[.3]{stateP}), (ii) velocity (\includeFGO[.3]{stateV}), (iii) IMU bias (\includeFGO[.3]{stateB}).

\subsection{Factors}
Each measurement introduces a residual that penalizes the deviation between predicted and observed quantities in appropriate Mahalanobis norms.

\noindent \textit{Prior factor (\includeFGO[.65]{fprior}).}
Anchors the first state and defines the reference frame:
\begin{equation}
J_{\text{prior}} = \|x_0 - \bar{x}_0\|_{\Sigma_0}^2.
\end{equation}

\noindent \textit{IMU preintegration factor (\includeFGO[.65]{fIMU}).}
Summarizes all inertial data between consecutive states~$k$ and~$k{+}1$, constraining relative motion in pose, velocity, and bias:
\begin{equation}
J_{\text{IMU},k} = 
\|r_{\text{IMU}}(x_k, x_{k+1})\|_{\Sigma_{\text{IMU}}}^2,
\label{eq:j_imu_prop}
\end{equation}
where $r_{\text{IMU}}$ is the preintegrated IMU residual (see \cite{forster_imu_2015,carlone_eliminating_2014} for more details on preintegration).

\noindent \textit{IMU bias random walk (\includeFGO[.65]{fRCB}).}
To ensure temporal consistency, biases are modeled as slow random walks:
\begin{equation}
J_{\text{bias},k} = 
\|b_a^{k+1}-b_a^{k}\|_{\Sigma_{ba}}^2 +
\|b_g^{k+1}-b_g^{k}\|_{\Sigma_{bg}}^2.
\label{eq:j_imu_bias}
\end{equation}

\noindent \textit{GNSS factor(\includeFGO[.65]{fPR-GPS}).}
Constrains the estimated position with the measured position provided by the GNSS receiver:
\begin{equation}
J_{\text{GNSS},m} =
\|p_k - p_k^{\text{GNSS}}\|_{\Sigma_{\text{GNSS}}}^2.
\label{eq:j_gnss}
\end{equation}



\subsection{Optimization objective}

Given the measurements and the preintegration factors, of different lengths, similar to \eqref{eq:FGO}, the complete optimization problem is obtained by combining all factors:
\begin{equation}
J(\mathcal{X}) =
J_{\text{prior}} +
\sum_k J_{\text{IMU},k} +
\sum_m J_{\text{GNSS},m},
\label{eq:FGO_nav}
\end{equation}
The GNSS measurements and the propagation of the inertial IMU data between consecutive states are modeled by equations \eqref{eq:j_imu_prop}-\eqref{eq:j_gnss} and echo the description given in \eqref{eq:FGO}.
Minimizing $J(\mathcal{X})$ yields the maximum a posteriori (MAP) trajectory consistent with all measurements, as the quantity in \eqref{eq:FGO_nav} can be interpreted to be the negative log-posterior conditional probability over the states, given the measurements.

\section{Proposed method}

In this section, we describe the proposed loosely coupled FGO architecture for GNSS/IMU measurement fusion. We start by describing the prior state-of-the-art work and then extend it to the real-time scenario by enabling smoothing latency and marginalization techniques.

\subsection{Current state-of-the-art: SFGO}

The Standard FGO (SFGO) \cite{11028340} approach is a post-processing method that constructs the factor graph from batch-collected GNSS/IMU measurements and subsequently applies an iterative nonlinear least-squares algorithm, implemented via GTSAM \cite{gtsam}, to obtain a smooth trajectory. Because the FGO problem is updated only when new GNSS measurements are available (as illustrated in Fig.~\ref{fig:fgo_scheme}), SFGO provides optimal trajectory estimates exclusively at those epochs. From a real-time perspective, this approach presents two main limitations: (i) the absence of real-time output and (ii) reduced service availability, as position estimates are produced only when new GNSS observations are available - which, in dense urban environments, can be infrequent.

\subsection{The proposed method: RTFGO}

Building upon SFGO, we propose several modifications to enable real-time capable fusion. 
First, we incorporate IMU-only propagation during GNSS outages to increase service availability. This trades off positioning accuracy, as prolonged propagation amplifies IMU bias drift, in general. Therefore, the duration of consecutive IMU propagation steps is limited by a time threshold dependent on the IMU performance. Second, when new GNSS measurements become available, the factor graph is expanded, and optimization is performed using the iSAM2 algorithm~\cite{isam2} to update the solution. Then, the solution is provided immediately after optimization or with a controlled smoothing latency, allowing additional future measurements to refine past states (as discussed later). Finally, to maintain real-time performance, we apply fixed-lag marginalization, where states older than a specified lag are marginalized out of the graph, reducing the size of all subsequent optimization problems that are solved.


\subsection{RTFGO with smoothing latency}

To take advantage of the inherent smoothing capability of FGO, we analyze within the RTFGO framework the effect of introducing a smoothing latency, allowing future measurements to refine past or current estimates. {\color{black} In our implementation, the smoothing latency, denoted as $\tau$, represents the number of future GNSS measurements that are incorporated into the factor graph before producing the estimate of the current state. In other words, it represents a look-ahead window that delays the output so that additional measurements can refine the current and past states.}
In this scenario, IMU-only propagation is disabled, and results are evaluated only at timestamps where GNSS data are available, matching the availability of the SFGO and GNSS-only solutions. The marginalization lag is kept infinite, ensuring that no past states are removed from the graph.

The impact of the smoothing latency depends on the quality of the future measurements, adding to the FGO formulation new factors that improve or worsen the localization results. Accuracy performance improvement is not guaranteed. {\color{black} A smoothing latency equal to the length of the full trajectory (i.e., the maximum possible value) results in a batch-processing method similar to SFGO.}

\subsection{RTFGO with marginalization lag}

For all previously proposed FGO solutions, the factor graph grows unbounded with the number of measurements, and thus the running time of the FGO solver can exceed imposed real-time constraints.

Marginalizing out states older than a fixed lag bounds the use of memory and computation resources, but it also removes factors that could otherwise recondition the trajectory for better results. In FGO terms, eliminating past variables replaces their connected factors with a condensed prior fixed at the current linearization point. As a result, information useful for bias correction is lost. Therefore, the solution is more sensitive to local noise/outliers and we reduce the capability of FGO to refine earlier states as new, more accurate measurements are available (in the scenario where smoothing latency is also used).


\section{Experimental Results}

In the following, we present numerical results for RTFGO and comparisons with the state of the art.
Experiments were performed on a MacBook Air~M3, and computation times were recorded using \texttt{perf\_counter} from Python's \texttt{time} library.
Our code and data are freely available at
\url{https://codeberg.org/3T-NAFGO/RealTimeLocalizationFGO}.

\subsection{The dataset}

The experimental evaluation uses the \emph{UrbanNav-HK-Medium-Urban-1} dataset, collected on the 17$^\text{th}$ of May 2021 between 02:33:31 and 02:46:15 (GPST). The data was recorded in a dense urban canyon inside Hong Kong \cite{Hsunavi.602}. 

\begin{figure*}[!ht]
    \subfloat[Ground truth and GNSS (Loops 1 and 2)]{\label{fig:loops_gt}\includegraphics[trim={0 0 0 30}, clip, width=.67\columnwidth]{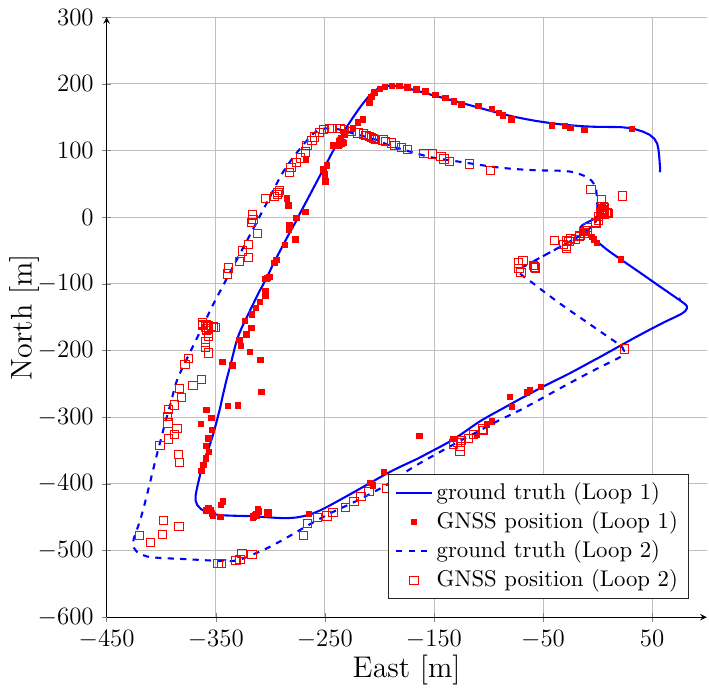}}\hfill
    \subfloat[Estimated trajectories (Loop 1)]{\label{fig:loops_1}\includegraphics[trim={0 0 0 30}, clip, width=.67\columnwidth]{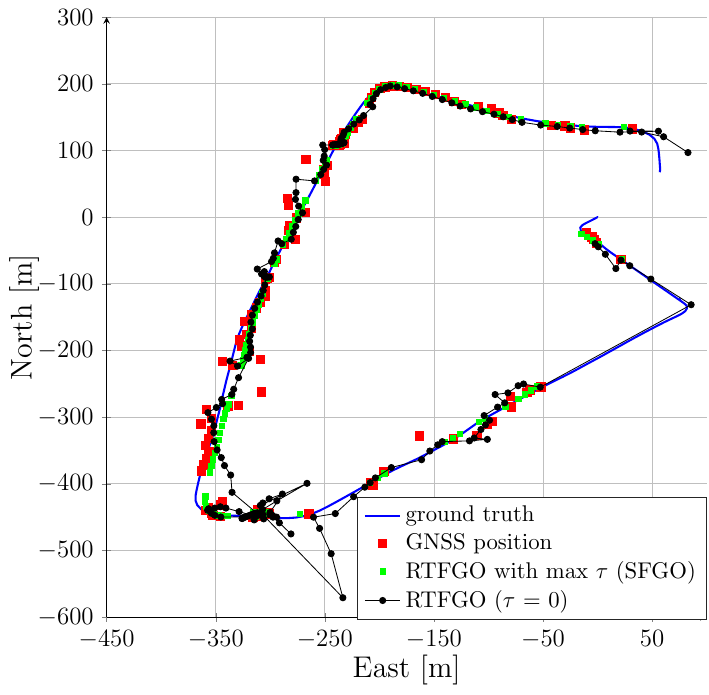}}\hfill
    \subfloat[Estimated trajectories (Loop 2)]{\label{fig:loops_2}\includegraphics[trim={0 0 0 30}, clip, width=.67\columnwidth]{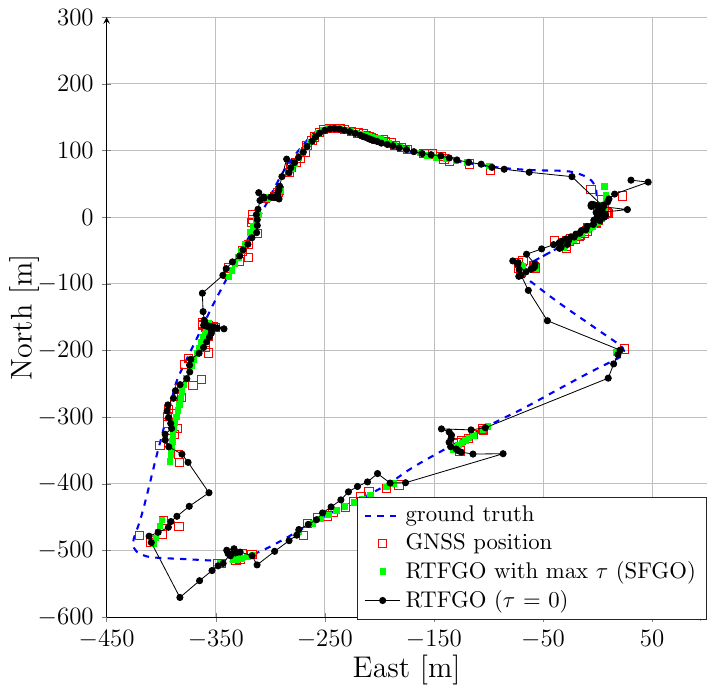}}
    \caption{Comparison of SFGO and RTFGO performance against GNSS-only and ground truth data (Loops 1 and 2)}
\end{figure*}

Two complete loops are performed over the same path within this time interval, hereafter referred to as \emph{Loop~1} and \emph{Loop~2}, with summary details provided in Table~\ref{tab:dataset-segments}, and their corresponding trajectories shown in Fig.~\ref{fig:loops_gt}.
\begin{table}[h]
\centering
\caption{Dataset loops and GNSS availability.}
\begin{tabular}{lcccc}
\toprule
Segment & Start time (GPST) & Duration [s] & GNSS Availability [\%] \\
\midrule
Loop~1 & 02:33:31 & 336 & 42.0 \\
Loop~2 & 02:39:13 & 422 & 38.9 \\
\bottomrule
\end{tabular}
\label{tab:dataset-segments}
\end{table}

The RTKLIB library \cite{RTKLIB} is used to compute the GNSS-only position from the raw GNSS measurements. The relevant RTKLIB parameters are: the elevation mask, which was set to $15^\circ$, the tropospheric model, which is the SBAS MOPS model, and the ionospheric model, which uses the Klobuchar model (the broadcast ionosphere option). All data streams are aligned to GPS~time (GPST).

The urban characteristics of the tracks are observed from the GNSS availability, which was only approximately 40\% in both loops. Moreover, we expect the GNSS positioning errors to be significantly larger than in open-sky conditions, where 3D Root Mean Squared Error (RMSE) values typically stay in the range of 5 m to 10 m. 

\subsection{Evaluation metrics}


The estimates computed by the proposed method are aligned with the ground-truth (GT) timestamps. If GT data are unavailable at specific timestamps (as in propagation-only states), we perform linear interpolation. This approach is suitable for the evaluation of real-time accuracy, particularly since GNSS provides real-time measurements but with limited availability.


The aforementioned metric is appropriate for our solution and for comparison with GNSS-only results, as both provide real-time outputs, unlike the previously proposed SFGO \cite{11028340}, which operates as a batch optimization method.

This methodology enables the analysis of solution availability with respect to the GT time grid, which spans 764~s at a rate of 1~Hz. The availability plots illustrate the number of samples (epochs in GNSS parlance) whose positioning error remains below a specified threshold. We define the \emph{service availability}~$A(\theta)$ as the fraction of all epochs - including those during GNSS outages - for which a valid position solution exists and the 3D position RMSE does not exceed a threshold~$\theta$. Formally,
\begin{equation}
A(\theta) = 
\frac{N_{\text{valid,\,error} \le \theta}}{N_{\text{total}}} \times 100~[\%],
\end{equation}
where $N_{\text{valid,\,error} \le \theta}$ is the number of samples with a valid
solution whose 3D position RMSE is below~$\theta$, and $N_{\text{total}}$ is the
total number of samples in the ground-truth trajectory.
Thus, $A(\theta)$ combines both positioning accuracy and continuity into a single
metric: outages reduce the achievable maximum availability, while higher
accuracy shifts the curve to the left, toward smaller~$\theta$.

For all the evaluated methods, the IMU noise parameters are exactly as provided in the dataset, and the GNSS position covariances are taken from the RTKLIB solution and scaled by a factor of two.

\subsection{Results}

Since SFGO currently serves as the baseline for loosely coupled GNSS/IMU fusion using FGO, we compare against this benchmark on both loops. The estimated positioning trajectories, for both loops using different methods, are shown in Fig.~\ref{fig:loops_1} and \ref{fig:loops_2}. We show the GT, the GNSS-only solution, the batch RTFGO result (equivalent to SFGO), and the proposed RTFGO method {\color{black} configured for real-time output with a maximum of 4 s of IMU-only propagation}. Performance is greatly affected by the severe GNSS degradation in urban canyons with large multipath interference, as expected.

As shown in Table~\ref{tab:accuracy}, {\color{black} RTFGO matches the accuracy of SFGO when configured with maximum smoothing latency and improves upon the GNSS-only solution. However, without IMU-only propagation, it provides no gain in service availability and behaves essentially as a post-processing method, just as SFGO does. We report the result of SFGO provided by \cite{11028340} only for Loop 2, as this is the only accuracy provided by the authors and no source code is available for reproducibility.}

\begin{figure}[h!]
    \centering
    \includegraphics[width=.8\columnwidth]{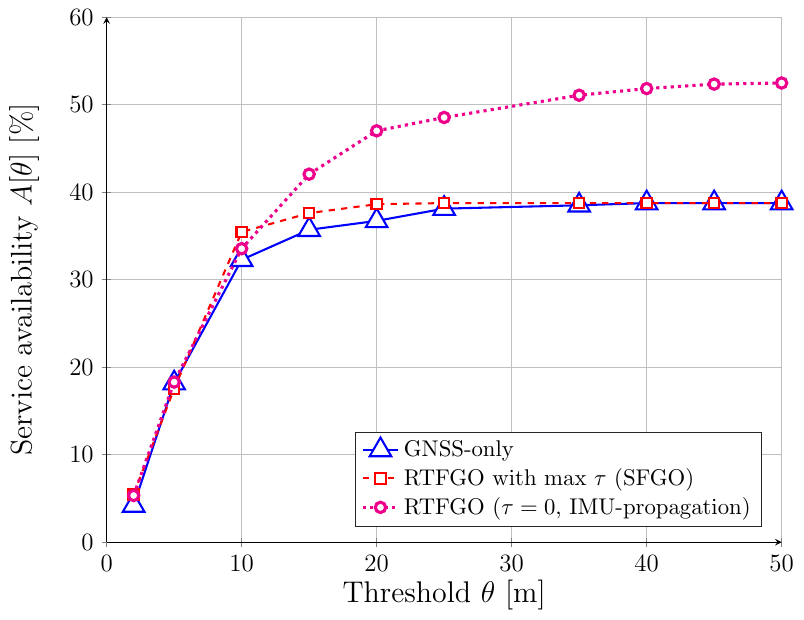}
    \caption{Service availability of the proposed methods}
    \label{fig:service_availability}
\end{figure}

{\color{black} By disabling smoothing latency, i.e., $\tau = 0$, and enforcing real-time operation, RTFGO substantially increases service availability, as shown in Fig.~\ref{fig:service_availability}. This comes at the cost of reduced accuracy (see Table~\ref{tab:accuracy}), since future measurements no longer refine past states and IMU-only propagation accumulates bias-induced drift.}


A particular difficulty {\color{black} of the real-time operation} is the cold start, when no reliable initial heading is available. Since yaw cannot be observed from the IMU alone, several consecutive GNSS fixes (typically 3–5) are required before the first proper result is reported. This affects the localization service availability at the start.

\begin{figure}[h!]
    \centering
    \includegraphics[width=.9\columnwidth]{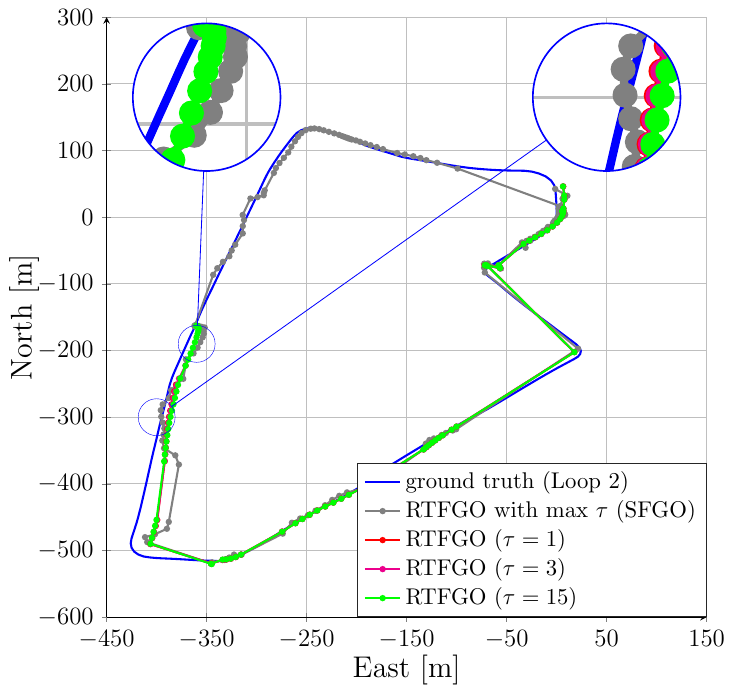}
    \caption{Smoothing latency impact on the RTFGO trajectory estimated for Loop 2.}
    \label{fig:smoothing_latency_trajectory}
\end{figure}

{\color{black} Increasing smoothing latency can improve accuracy when real-time constraints are removed, but the effect is two-sided: low-quality past measurements may worsen earlier estimates or provide no improvement.} To highlight this point, we select two segments on the track of Loop 2 to present them in opposition (see Fig. \ref{fig:smoothing_latency_trajectory}). For each segment, we arbitrarily select a state and plot the positioning error against the smoothing latency and the optimization cost objective provided by the iSAM2 algorithm (obtained by summing the costs related to all the factors connected to that state).

\begin{figure}[!t]
  \centering
  \subfloat[Position error evolution versus smoothing latency $\tau$ for the position at (02:43:35.000 GPST)]{%
    \includegraphics[width=.9\columnwidth]{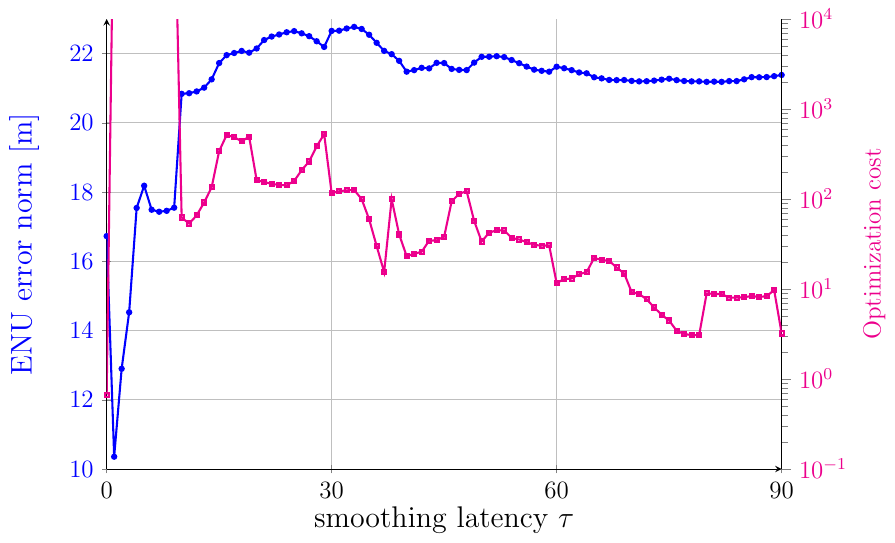}\label{fig:a}}
  \hfill
  \subfloat[Position error evolution versus smoothing latency $\tau$ for the position at (02:43:48.000 GPST)]{%
    \includegraphics[width=.9\columnwidth]{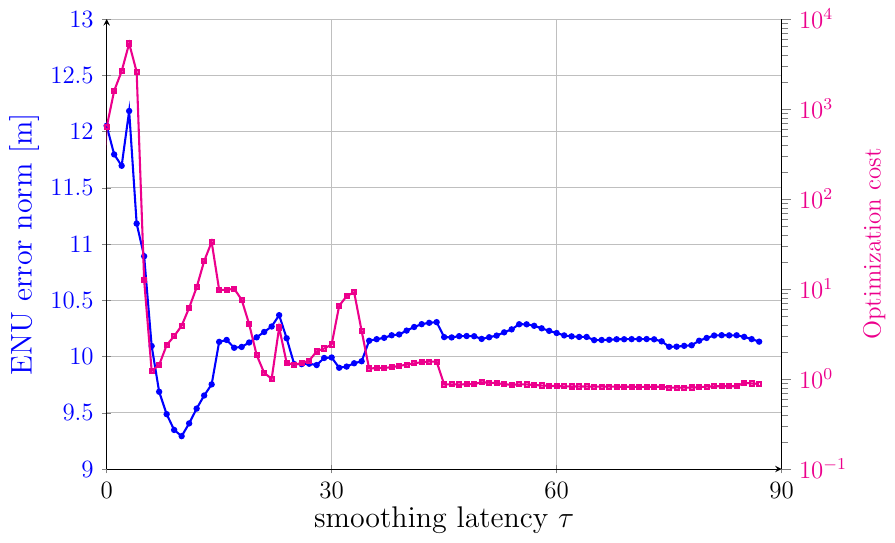}\label{fig:b}}
  \caption{Position error and optimization cost evolution.}
  \label{fig:state_evolution_smoothing_latency}
\end{figure}

\begin{figure}[t!]
    \centering
    \includegraphics[width=.9\columnwidth]{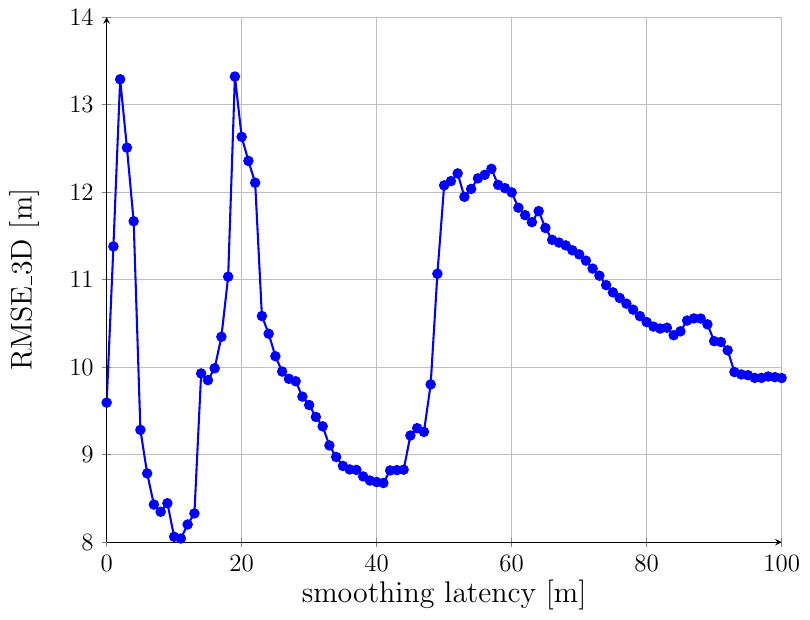}
    \caption{3D RMSE versus smoothing lag on the RTFGO trajectory for Loop 2.}
    \label{fig:enu_stats_smoothing_loop_2}
\end{figure}

In both cases, the optimization cost decreases as additional measurements are introduced, confirming that the optimizer effectively incorporates the new information into the estimation process. However, for the segment affected by degraded GNSS data, the positioning error increases with larger smoothing latency, indicating that the added measurements introduce inconsistent or biased constraints (see Fig~\ref{fig:state_evolution_smoothing_latency}). The same effect is observed in Fig. \ref{fig:enu_stats_smoothing_loop_2} for the 3D RMSE of the Loop 2 trajectory as a function of the smoothing latency $\tau$. 

\begin{figure}[t!]
    \centering
    \includegraphics[width=.9\columnwidth]{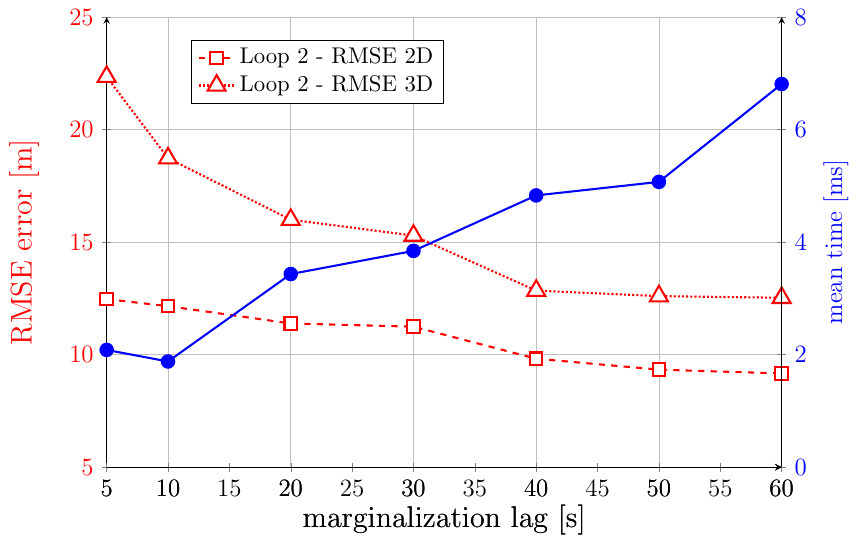}
    \caption{Marginalization lag impact on the RTFGO trajectory estimated for Loop 2.}
    \label{fig:marginalization_error_mean_latency}
\end{figure}

{\color{black} Finally,} we analyze the effect of the marginalization lag within the RTFGO framework. For simplicity, and to isolate the effect of the marginalization lag, we disable the smoothing latency and the IMU propagation. Experiments are performed only on Loop 2, and for various marginalization lags, the resulting positioning errors and mean optimization times are presented in Fig.~\ref{fig:marginalization_error_mean_latency}.

As expected, the positioning errors decrease as the lag increases, reaching 12.6 m on average for the 3D RMSE at the marginalization lag of 50 s and 5 ms optimization computation time, clearly highlighting the trade-off between marginalization lag, accuracy, and optimization computational effort. For real-time operation, our target is to provide the best possible localization results, using a recent buffer of GNSS/IMU measurements, at the millisecond timescale.


\begin{table}[t!]
\centering
\caption{3D RMSE accuracy for Loop~1 and Loop~2}
\begin{tabular}{l p{2cm} r r}
\toprule
Method & Configuration &
\multicolumn{1}{c}{\begin{tabular}{@{}c@{}}Loop~1\\RMSE\_3D [m]\end{tabular}} &
\multicolumn{1}{c}{\begin{tabular}{@{}c@{}}Loop~2\\RMSE\_3D [m]\end{tabular}} \\
\midrule
GNSS-only & via RTKLIB & 29.20 & 13.24 \\
SFGO & \cite{11028340} & --- & 9.89 \\
RTFGO & Batch (max $\tau$)& 27.91 & 9.33 \\
RTFGO & RT ($\tau$ = 0)& 33.31 & 11.83 \\
\bottomrule
\end{tabular}
\label{tab:accuracy}
\end{table}


\section{Conclusions}

We introduce RTFGO, a real-time capable GNSS/IMU fusion method based on Factor Graph Optimization (FGO), and benchmark it on the UrbanNav-HK-MediumUrban-1 dataset. RTFGO matches the performance of previous state-of-the-art batch GNSS/IMU FGO approaches. Furthermore, it enables real-time measurement fusion and allows the analysis of the trade-offs introduced by smoothing latencies and marginalization lags. The results show how positioning accuracy can be traded for improved service availability and real-time performance in challenging urban environments.


Future work will focus on extending RTFGO to a tightly coupled GNSS/IMU integration, which is expected to improve both accuracy and robustness to outliers. In addition, the increased modularity of FGO allows the inclusion of additional dynamical or contextual constraints. We plan to exploit this by incorporating environment-aware factors during GNSS outages, such as zero-velocity updates, map-based height priors, and road geometry constraints.

This work has been submitted to the IEEE for possible publication. Copyright may be transferred without notice, after which this version may no longer be accessible

\end{document}

%% file: tikz_styles.tex
\tikzstyle{valve}=[thick,decoration={markings,  
  mark connection node=dmp,
  mark=at position 0.5 with 
  {
    \node (dmp) [thick,inner sep=0pt,transform shape,rotate=0,minimum width=15pt,minimum height=10pt,draw=none] {};
\draw[thick, fill=white](dmp.north west)--(dmp.south east)--(dmp.north east)--(dmp.south west)--(dmp.north west);
\draw[thick](dmp.center)--++(0,8pt)coordinate(a);
\draw[thick, fill=black] ([shift=(0:6pt)]a) arc (0:180:6pt);
  }
}, decorate]

\tikzstyle{spring}=[thick,decorate,decoration={zigzag,pre length=10,post length=10,segment length=10}]

\tikzstyle{damper}=[thick,decoration={markings,  
  mark connection node=dmp,
  mark=at position 0.5 with 
  {
    \node (dmp) [thick,inner sep=0pt,transform shape,rotate=-90,minimum width=15pt,minimum height=3pt,draw=none] {};
    \draw [thick] ($(dmp.north east)+(2pt,0)$) -- (dmp.south east) -- (dmp.south west) -- ($(dmp.north west)+(2pt,0)$);
    \draw [thick] ($(dmp.north)+(0,-5pt)$) -- ($(dmp.north)+(0,5pt)$);
  }
}, decorate]
\tikzstyle{ground}=[fill,pattern=my grid,draw=none,minimum width=0.75cm,minimum height=0.3cm]

\tikzstyle{box}=[rounded corners=3, minimum height=.75cm, minimum width=1.5cm,draw, semithick, fill=gray!40]
\tikzstyle{gain}=[draw,semithick,isosceles triangle,inner sep=1pt, fill=gray!40]
\tikzstyle{ball}=[fill=gray,draw=black,circle,draw,minimum height=.75cm]
\tikzstyle{arrow}=[ultra thick,-latex]
\tikzstyle{arrowtr}=[,blue, semithick,decoration={markings,mark=at position 0.5 with {\arrow[scale=.7]{triangle 60}}},postaction={decorate}]
\tikzstyle{arrowtrr}=[,blue, semithick,decoration={markings,mark=at position 0.5 with {\arrow[scale=.7]{triangle 60 reversed}}},postaction={decorate}]
\tikzstyle{sum}=[circle,draw,minimum height=8]
\tikzstyle{block}=[rounded corners=3, minimum height=.5cm, minimum width=1cm,draw, semithick,fill=none]

\pgfdeclarepatternformonly[\GridSize]{my grid}{\pgfqpoint{-1pt}{-1pt}}{\pgfqpoint{\GridSize}{\GridSize}}{\pgfqpoint{\GridSize}{\GridSize}}%
{
     \pgfsetlinewidth{0.25pt}
     \pgfpathmoveto{\pgfqpoint{0pt}{0pt}}
     \pgfpathlineto{\pgfqpoint{\GridSize}{\GridSize + 0.1pt}}
     \pgfpathmoveto{\pgfqpoint{0pt}{0pt}}
     \pgfusepath{stroke}
}

\newdimen\GridSize
\tikzset{
    GridSize/.code={\GridSize=#1},
    GridSize=20pt
}

\tikzstyle{normalt}=[font=\small, fill=white,draw=black]
\tikzstyle{linet1}=[color=blue,thick,solid]
\tikzstyle{linet2}=[color=blue,dashed,thick]
\tikzstyle{linet3}=[color=blue,dotted,very thick]
\tikzstyle{linet4}=[color=blue,dashdotted, thick]
\tikzstyle{combt}=[ycomb,color=blue,solid,mark=*,mark options={solid}]

\tikzstyle{axa}=[thin,->]
\tikzstyle{fparrw}=[very thin,-latex',shorten >=.5pt]
\tikzstyle{bfparrw}=[very thin,latex'-latex',shorten <=.5pt,shorten
  >=.5pt]
  
\pgfplotsset{compat=newest}
\pgfplotsset{every tick label/.append style={font=\Large}}
\pgfplotsset{legend style={font=\small}}
\pgfplotsset{
		x label style={at={(axis description cs:0.5,-0.05)},anchor=north, font=\small},
    y label style={at={(axis description cs:-0.05,.5)},rotate=0,anchor=south, font=\small},
  }
\pgfplotsset{every tick label/.append style={font=\normalfont}}
\pgfplotsset{every axis x label/.append style={at={(current axis.south)},below=5mm, font=\large}}
\pgfplotsset{every axis y label/.append style={at={(current axis.west)},yshift=10mm, font=\large}}

\tikzset{
semilog lines/.style={thin, gray},
semilog lines 2/.style={semilog lines,
gray!50 },
semilog half lines/.style={semilog lines 2,
dotted },
semilog label x/.style={semilog lines,
below,font=\small, black},
semilog label y/.style={semilog lines,
right,font=\small,black},
asymp lines/.style={thin, densely dashed,black},
Bode lines/.style={semithick, black}
}
\tikzset{
Nyquist lines/.style={semithick, black},
Nyquist grid/.style={thin,gray},
Nyquist label axes/.style={Nyquist grid,font=\small},
Nyquist label points/.style={font=\small}
}

\tikzset{
        hatch distance/.store in=\hatchdistance,
        hatch distance=10pt,
        hatch thickness/.store in=\hatchthickness,
        hatch thickness=2pt
    }

    \makeatletter
    \pgfdeclarepatternformonly[\hatchdistance,\hatchthickness]{mypattern}
    {\pgfqpoint{0pt}{0pt}}
    {\pgfqpoint{\hatchdistance}{\hatchdistance}}
    {\pgfpoint{\hatchdistance-1pt}{\hatchdistance-1pt}}%
    {
        \pgfsetcolor{green!40}
        \pgfsetlinewidth{\hatchthickness}
        \pgfpathmoveto{\pgfqpoint{0pt}{0pt}}
        \pgfpathlineto{\pgfqpoint{\hatchdistance}{\hatchdistance}}
        \pgfusepath{stroke}
    }
    \pgfdeclarepatternformonly[\hatchdistance,\hatchthickness]{mypattern2}
    {\pgfqpoint{0pt}{0pt}}
    {\pgfqpoint{\hatchdistance}{\hatchdistance}}
    {\pgfpoint{\hatchdistance-1pt}{\hatchdistance-1pt}}%
    {
        \pgfsetcolor{red!40}
        \pgfsetlinewidth{\hatchthickness}
        \pgfpathmoveto{\pgfqpoint{0pt}{0pt}}
        \pgfpathlineto{\pgfqpoint{\hatchdistance}{\hatchdistance}}
        \pgfusepath{stroke}
    }
    \pgfdeclarepatternformonly[\hatchdistance,\hatchthickness]{mypattern3}
    {\pgfqpoint{0pt}{0pt}}
    {\pgfqpoint{\hatchdistance}{\hatchdistance}}
    {\pgfpoint{\hatchdistance-1pt}{\hatchdistance-1pt}}%
    {
        \pgfsetcolor{blue!40}
        \pgfsetlinewidth{\hatchthickness}
        \pgfpathmoveto{\pgfqpoint{0pt}{0pt}}
        \pgfpathlineto{\pgfqpoint{\hatchdistance}{\hatchdistance}}
        \pgfusepath{stroke}
    }

    \makeatletter
\tikzset{hatch distance/.store in=\hatchdistance,hatch distance=5pt,hatch thickness/.store in=\hatchthickness,hatch thickness=5pt}

\pgfdeclarepatternformonly[\hatchdistance,\hatchthickness]{north east hatch}
    {\pgfqpoint{-\hatchthickness}{-\hatchthickness}}
    {\pgfqpoint{\hatchdistance+\hatchthickness}{\hatchdistance+\hatchthickness}}
    {\pgfpoint{\hatchdistance}{\hatchdistance}}%
    {
        \pgfsetcolor{\tikz@pattern@color}
        \pgfsetlinewidth{\hatchthickness}
        \pgfpathmoveto{\pgfqpoint{-\hatchthickness}{-\hatchthickness}}       
        \pgfpathlineto{\pgfqpoint{\hatchdistance+\hatchthickness}{\hatchdistance+\hatchthickness}}
        \pgfusepath{stroke}
    }
\pgfdeclarepatternformonly[\hatchdistance,\hatchthickness]{north west hatch}
    {\pgfqpoint{-\hatchthickness}{-\hatchthickness}}
    {\pgfqpoint{\hatchdistance+\hatchthickness}{\hatchdistance+\hatchthickness}}
    {\pgfpoint{\hatchdistance}{\hatchdistance}}%
    {
        \pgfsetcolor{\tikz@pattern@color}
        \pgfsetlinewidth{\hatchthickness}
        \pgfpathmoveto{\pgfqpoint{\hatchdistance+\hatchthickness}{-\hatchthickness}}
        \pgfpathlineto{\pgfqpoint{-\hatchthickness}{\hatchdistance+\hatchthickness}}
        \pgfusepath{stroke}
    }

\pgfdeclarepattern{
  name=hatch,
  parameters={\hatchsize,\hatchangle,\hatchlinewidth},
  bottom left={\pgfpoint{-.1pt}{-.1pt}},
  top right={\pgfpoint{\hatchsize+.1pt}{\hatchsize+.1pt}},
  tile size={\pgfpoint{\hatchsize}{\hatchsize}},
  tile transformation={\pgftransformrotate{\hatchangle}},
  code={
    \pgfsetlinewidth{\hatchlinewidth}
    \pgfpathmoveto{\pgfpoint{-.1pt}{-.1pt}}
    \pgfpathlineto{\pgfpoint{\hatchsize+.1pt}{\hatchsize+.1pt}}
    \pgfpathmoveto{\pgfpoint{-.1pt}{\hatchsize+.1pt}}
    \pgfpathlineto{\pgfpoint{\hatchsize+.1pt}{-.1pt}}
    \pgfusepath{stroke}
  }
}

\pgfdeclarepattern{
  name=pLines,
  parameters={\hatchsize,\hatchangle,\hatchlinewidth},
  bottom left={\pgfpoint{-.1pt}{-.1pt}},
  top right={\pgfpoint{\hatchsize+.1pt}{\hatchsize+.1pt}},
  tile size={\pgfpoint{\hatchsize}{\hatchsize}},
  tile transformation={\pgftransformrotate{\hatchangle}},
  code={
    \pgfsetlinewidth{\hatchlinewidth}
    \pgfpathmoveto{\pgfpoint{-.1pt}{-.1pt}}
    \pgfpathlineto{\pgfpoint{\hatchsize+.1pt}{\hatchsize+.1pt}}
    \pgfusepath{stroke}
  }
}

\tikzset{
  hatch size/.store in=\hatchsize,
  hatch angle/.store in=\hatchangle,
  hatch line width/.store in=\hatchlinewidth,
  hatch size=5pt,
  hatch angle=0pt,
  hatch line width=.5pt,
}

\tikzset{
  pLines size/.store in=\hatchsize,
  pLines angle/.store in=\hatchangle,
  pLines line width/.store in=\hatchlinewidth,
  pLines size=5pt,
  pLines angle=0pt,
  pLines line width=.5pt,
}

\tikzstyle{stateCB}=[draw=black, circle, fill=cyan!30, text=black, minimum width=3em, line width=1pt]
\tikzstyle{stateP}=[draw,Green, circle, fill=Green!40, text=black, minimum width=3em, line width=1pt]
\tikzstyle{stateV}=[draw,IndianRed, circle, fill=IndianRed!40, text=black, minimum width=3em, line width=1pt]
\tikzstyle{stateD}=[draw,DarkGoldenrod, circle, fill=DarkGoldenrod!40, text=black, minimum width=3em, line width=1pt]
\tikzstyle{stateB}=[draw,LightSlateBlue, circle, fill=LightSlateBlue!40, text=black, minimum width=3em, line width=1pt]

\tikzstyle{fRCB}=[draw=black, fill=red!20, text=black, minimum width=1.5em, minimum height=1.5em,preaction={fill, red!20}, pattern=hatch, pattern color=black, hatch size=7.5pt]
\tikzstyle{fIMU}=[draw=black, text=black, minimum width=1.5em, minimum height=1.5em, preaction={fill, green!10},pattern = dots]
\tikzstyle{fPR-GPS}=[draw,black, text=black, minimum width=1.5em, minimum height=1.5em,preaction={fill, blue!10},pattern=pLines, pattern color=blue, pLines size = 8pt, pLines angle=0]
\tikzstyle{fPR-Galileo}=[draw=black, text=black, minimum width=1.5em, minimum height=1.5em,preaction={fill,lightgray!20}, pattern=north west hatch,hatch distance=7pt,hatch thickness=.5pt,pattern color=black]
\tikzstyle{fDop-GPS}=[draw,black, text=black, minimum width=1.5em, minimum height=1.5em,preaction={fill, orange!10},pattern=pLines, pattern color=orange, pLines size = 8pt, pLines angle=0]
\tikzstyle{fDop-Galileo}=[draw=black, text=black, minimum width=1.5em, minimum height=1.5em,preaction={fill,LightSeaGreen!20}, pattern=north west hatch,hatch distance=7pt,hatch thickness=.5pt,pattern color=LightSeaGreen]
\tikzstyle{fprior}=[draw=black, text=black, minimum width=1.5em, minimum height=1.5em, preaction={fill, orange!30},pattern = bricks]    
\makeatother